# Learning via Long Short-Term Memory (LSTM) network for predicting strains in Railway Bridge members under train induced vibration

Amartya Dutta[1*], Kamaljyoti Nath[2]

[1]Indian Institute of Information Technology, Guwahati, India
*Corresponding author
amartyad7@gmail.com

[2]Indian Institute of Technology, Guwahati, India
kamaljyoti@iitg.ac.in

**Abstract.** Bridge health monitoring using machine learning tools has become an efficient and cost- effective approach in recent times. In the present study, strains in railway bridge member, available from a previous study conducted by IIT Guwahati has been utilized. These strain data were collected from an existing bridge while trains were passing over the bridge. LSTM is used to train the network and to predict strains in different members of the railway bridge. Actual field data has been used for the purpose of predicting strain in different members using strain data from a single member, yet it has been observed that they are quite agreeable to those of ground truth values. This is in-spite of the fact that a lot of noise existed in the data, thus showing the efficacy of LSTM in training and predicting even from noisy field data. This may easily open up the possibility of collecting data from the bridge with a much lesser number of sensors and predicting the strain data in other members through LSTM network.

## 1      Introduction

Strains in critical members of railway bridges under the action of moving train load are regularly monitored to understand the health of the structure. The strain in a bridge member can be a direct indicative whether the bridge is undergoing any possible degradation. The strains in bridge members are also monitored to appreciate the effect of any increase in axle load of the moving vehicle. However, any standard railway bridge comprises of a large number of members and a good number of these members are instrumented to acquire strain data while a train passes over the bridge. It is thus easily appreciated that the entire process of fixing strain gauges and collection of data is expensive as well as time consuming. Thus, an alternative strategy is considered to address this issue of strain data collection, which is very intricately associated with structural health monitoring.

Monitoring of bridge using machine learning has become very popular among researchers. Shu et al. [1] implemented a damage detection algorithm for railway bridge based on back propagation Artificial Neural Network (ANN) using the statistical properties of the dynamic responses of the structure as input for the ANN. Chalouhi et al. [2] presented a method that uses machine learning to detect and localize damage in railway bridges. They applied the strategy to a historical bridge and validated the proposed algorithm. The proposed method can be used to detect inconsistent responses that can be labelled as possible damage. Neves et al. [3] presented a model-free damage detection approach based on machine learning techniques. Artificial neural networks are trained with an unsupervised learning approach with input data in the form of accelerations gathered from the bridge. Malekjafarian et al. [4] proposed a new two-stage machine learning approach for bridge damage detection using the responses measured from sensors placed on a passing vehicle. Dutta [5] applied a density-based clustering technique on railway bridge strain data acquired from field to identify the data density and noisy data elements, which can be subsequently used in decision making for structural health monitoring. In recent years, LSTM has demonstrated noteworthy performance on a good number of real-world applications such as machine translation (Sutskever et al. [6]), speech recognition (Graves et al. [7]) and video classification (Ng et al., [8]). There has also been an increasing attention towards using LSTMs for time series prediction such as air pollution forecasting (Freeman et al. [9]), traffic flow prediction etc. (Tian and Pan [10]).

In the present study, learning via LSTM is proposed for predicting strains in railway bridge members utilizing already available strain time history of the same bridge for a good number of earlier conducted tests. The study presented here uses real data from the bridge and hence the study is more challenging due to the presence of noise and other associated uncertainty in data collection. Highly satisfactory agreement is observed between the predicted strain time history with ground truth for a good number of cases and hence provides an ideal opportunity to explore the application of such machine tools.

## 2    Description of Test Bridge and Data

The considered bridge is a twenty span truss bridge between Jalpaiguri Road and New Domohoni station under Alipurduar Division in West Bengal, India. The spans of the bridge are all equal and is 45.72 m carrying a single broad gauge track. The steel superstructures rest on concrete piers and abutments at two ends. One of the spans is taken for study and is shown in Fig. 1. A train with known axle loads (named here as Test train) is allowed to travel at different speeds. Strains of some critical members were measured. Some of the typical sensors attached to the bridge are shown also in Fig. 2. Further, data were also collected for passenger trains, whose axle loads are not exactly known. The data collection continued over four years and hence significant amount of measured data are available.

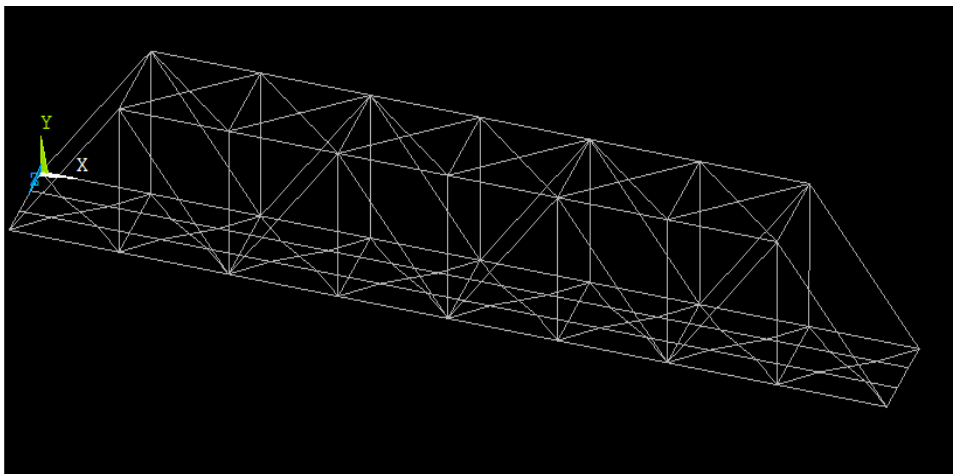

Fig. 1 One typical span of the railway truss bridge

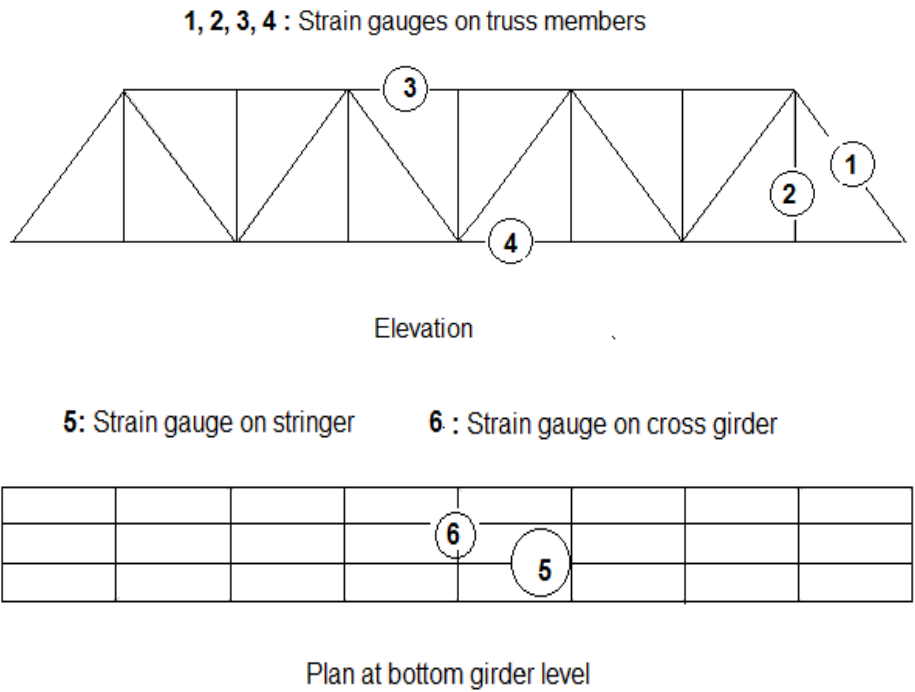

Fig. 2  A typical span of the bridge showing the sensor locations

Typical plots of measured strains are shown in Fig. 3. These are for different speed of test train as well as for service train. The data were collected at the sampling rate of 0.025 sec for train moving at different speeds. Thus, it may be observed that the amount of time series data in Fig. 3 are different as the train was made to move at different speeds. In the case of service train, the axle load for engine is much higher than that of the rest, which can also be observed from the initial higher values of strain. It may be appreciated that the strain in a particular member is correlated to strain in other members of the bridge on the same span. However, the measurement of all these strains at different locations are quite involved and expensive. Thus, the objective of the present study is to predict the strain in some of the critical bridge members using the strain data of any one of the members of the bridge.

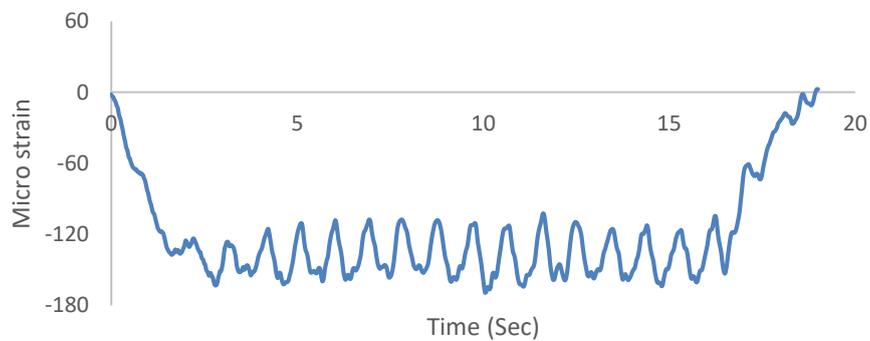

(a)  Location 1 (Test train at 50 kmph)

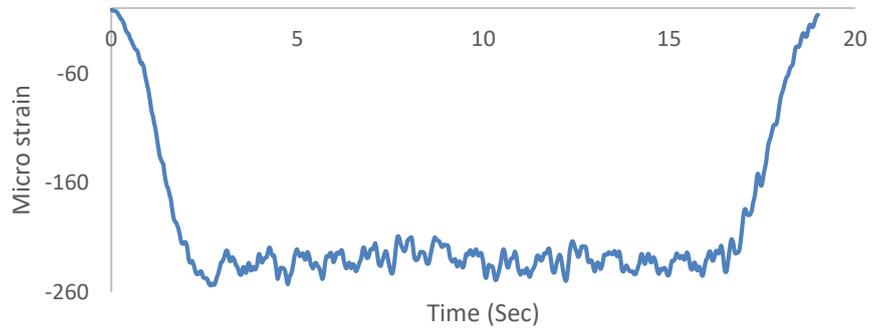

(b) Location 3 (Test train 50 kmph)

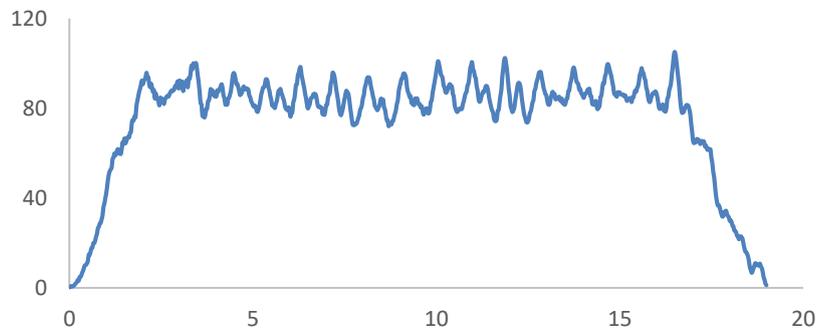

(c) Location 4 (Test train at 50 kmph)

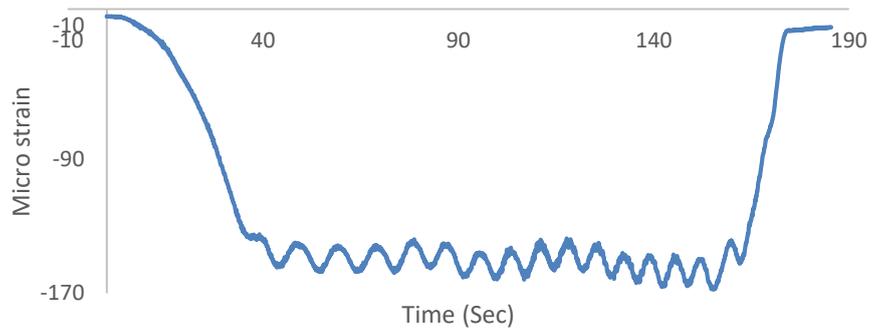

(d) Location 1 (Test train at 5kmph)

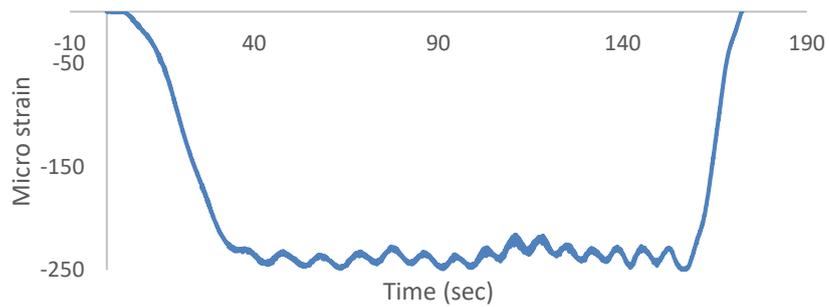

(e) Location 3 (Test train at 5kmph)

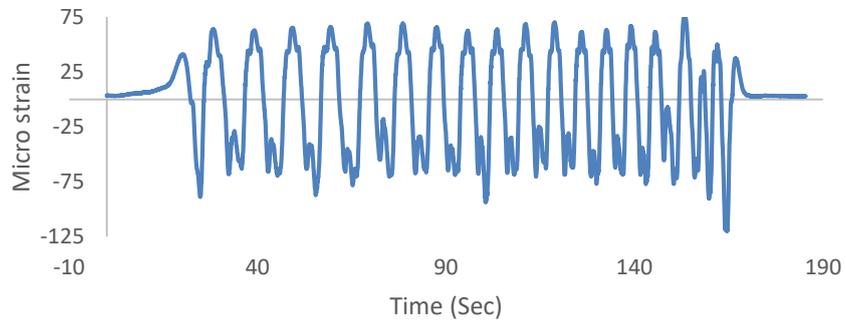

(f) Location 5 (Test train at 5kmph)

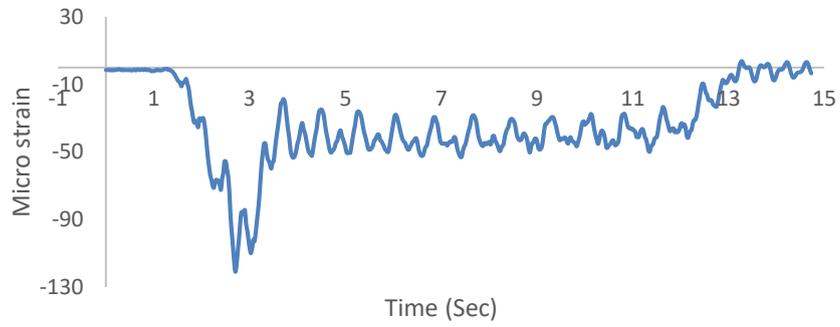

(g) Location 1 (Passenger train at 5kmph)

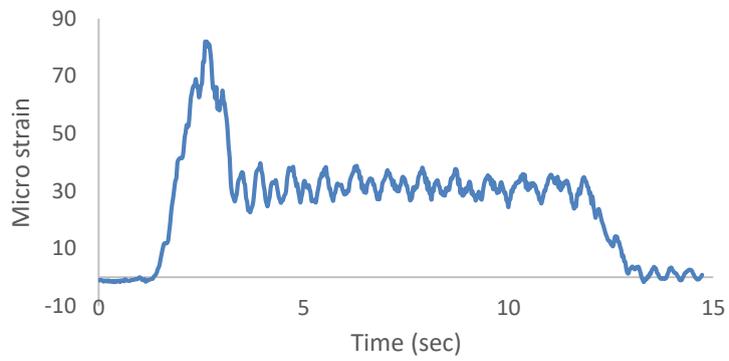

(h) Location 4 (Passenger train at 5kmph)

Fig. 3: Typical plots of strain time histories corresponding to different sensor locations, speed and train

## 3    LSTM Network

RNNs are a generalization of feedforward neural networks, one that allows the neural network also to learn by remembering the past information. However, in case of long-term dependencies, RNNs face vanishing and exploding gradient problems. That is why a gated variant of the RNN, termed as LSTM has been used for the work in this paper. The gates in these units are used to control the flow of information that passes through the state of the cell. The chosen type of gated cell (LSTM) was introduced two decades ago [11] and has now gained popularity in the context of language modeling. However, the work done in the present paper attempts to exploit the advantage of LSTMs when it comes to time series forecasting. As a result, an attempt has been made to predict the various forms of strain the bridge members would experience using the time series data. The formulation of LSTM cells is as defined by [11,12]. Assuming $i_t$, $f_t$, $o_t$, $c_t$ and $h_t$ to indicate the values of the input gate, forget gate, output gate, memory cell and hidden state at time $t$ in the sequence respectively and $x_t$ be the input of the system at time $t$, the architecture of the LSTM cell can be defined as follows

$$i_t = \sigma(W_{xi}x_t + W_{hi}h_{t-1} + W_{ci}c_{t-1} + b_i) \tag{1}$$
$$f_t = \sigma(W_{xf}x_t + W_{hf}h_{t-1} + W_{cf}c_{t-1} + b_f) \tag{2}$$
$$c_t = f_t \odot c_{t-1} + i_t \odot \tanh(W_{xc}x_t + W_{hc}h_{t-1} + b_c) \tag{3}$$
$$o_t = \sigma(W_{xo}x_t + W_{ho}h_{t-1} + W_{co}c_t + b_o) \tag{4}$$
$$h_t = o_t \odot \tanh(c_t) \tag{5}$$

## 4    Time Series Forecasting using LSTM

The strain data collected from different bridge members contains multiple features that vary over time. The work done in this paper uses LSTM to learn from the time series data. Since the aim is, given a time varying feature as the input to the model, the strains in other members are to be predicted, the time series data is converted such that it resembles supervised learning. Therefore, the data is transformed into a sliding window format such that every sample of the time series data is of the form
Input: [$x_i$; $x_{i+1}$ ;... $x_{i+T-1}$]
Output: $y_{i+T-1}$
Where $T$ is the sequence length, $x_i$ is the input feature and $y_i$ is the feature to be predicted. Following this, the time series model is used for the desired forecasting task. The strain corresponding to a location (taken as 1 in the present study) acts as our input, while the targets are strains at different locations considered one at a time that vary according to the cases as explained in the next section.

## 5    Prediction of strain using LSTM

A number of cases of prediction of strains are studied to understand the LSTM and its efficiency for each of the cases. The training is time consuming as different associated parameters are to be tuned to achieve the best possible prediction using LSTM. Parameters adopted for different cases are mentioned below the plots of each of the cases. In order to evaluate the accuracy of prediction, Root Mean Square Error (RMSE) is calculated between the predicted and target strain time history. While a smaller value of RMSE is indicative of a better performance in prediction, the magnitude of RMSE alone does not very clearly convey the extent of perfection achieved in such prediction. In view of this, a ratio is calculated for $L_2$ norm of error vector between Target and Predicted strain time history to $L_2$ norm of Target strain time history. If this ratio is deducted from unity and expressed in percentage, provides a measure of accuracy, which is physically interpretable. In the present study, both RMSE and % accuracy are evaluated for all the cases as detailed below.

### 5.1 Prediction Case-1

The first problem is the prediction of strain time history corresponding to location 3 by training a LSTM model with strain time history corresponding to location 1. The training and predictions are carried out with strain measured from the bridge due to the passage of test train at 60 kmph. The predicted strain time history and the target strain time history measured in field are shown in Fig. 4. Very good agreement is observed between the predicted strain time history and the ground truth, which is the available measured strain time history corresponding to the same location 3 of the existing bridge.

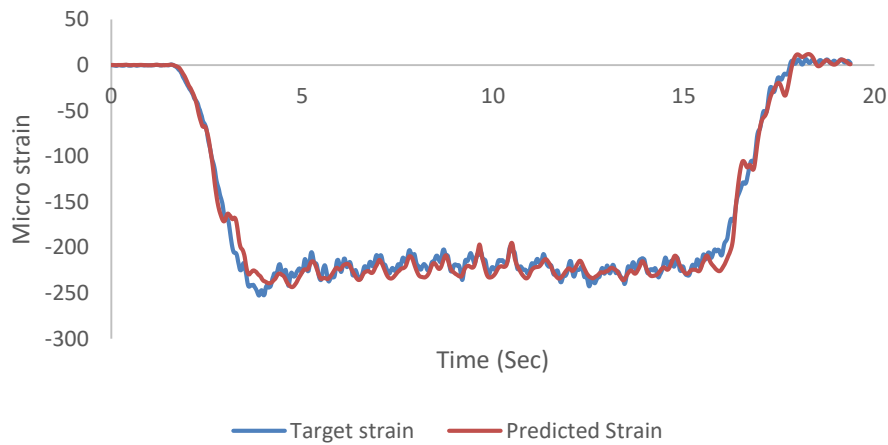

Fig. 4 Target and predicted strain time history at location 3 for Test train at 50 kmph (20 LSTM neurons, 1 hidden layer with 30 neurons, window size=50)

### 5.2 Prediction Case-2

The second problem is similar to the first case, where the training and prediction corresponds to the test train moving at a velocity of 5 kmph. The prediction of strain time history corresponding to location 3 is done by training a LSTM model with strain time history of location 1. Since the train is moving at a much lesser speed, the data size is quite different in this case as may be seen from Fig. 2 (d-e). The predicted strain time history and the target strain time history (ground truth) are shown in Fig. 5. Very good agreements are observed in this case as well.

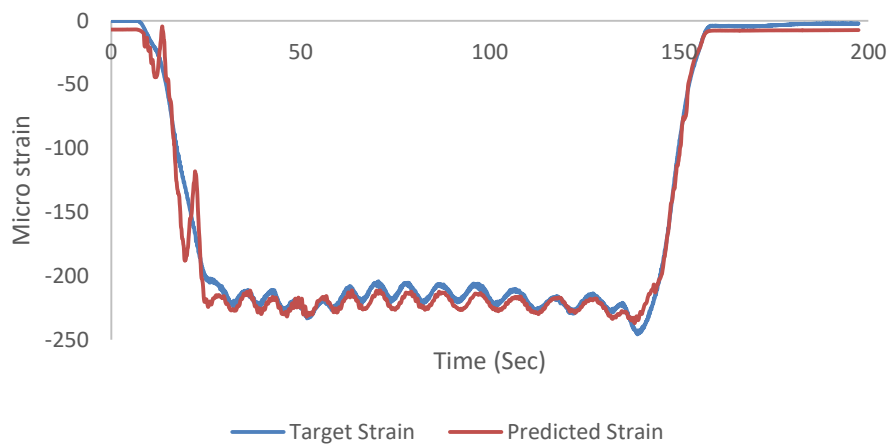

Fig. 5 Target and predicted strain time history at location 3 for Test train at 5 kmph (10 LSTM neurons, 1 hidden layer with 30 neurons, window size=50)

### 5.3 Prediction Case-3

As may be seen from Fig. 3(c, f), the patterns of strain time history corresponding to locations 4 and 5 are quite different than what have been observed corresponding to locations 1 and 3. In order to appreciate the applicability of LSTM, it is tried to predict strain time history corresponding to location 4 and 5 using the strain time history data corresponding to location 1. These are shown in Fig. 6 and 7 respectively. In this cases as well, very good agreements are observed between predicted strain time history and ground truth for the respective cases.

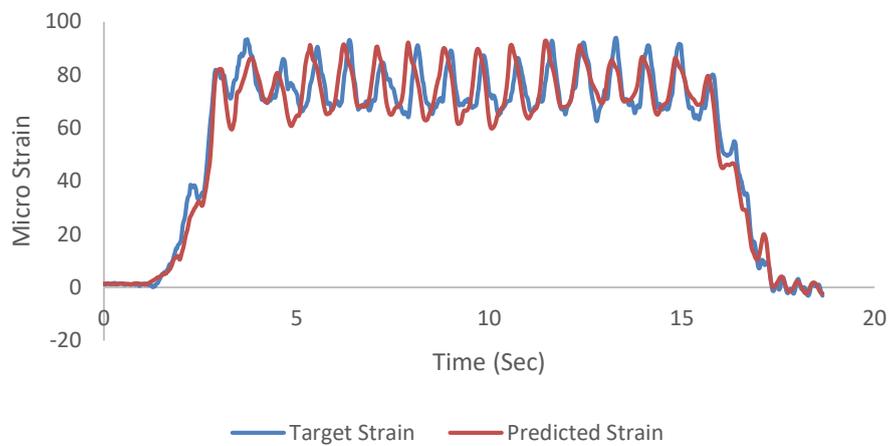

Fig. 6 Target and predicted strain time history at location 4 for Test train at 50 kmph (20 LSTM neurons, 1 hidden layer with 50 neurons, window size=50)

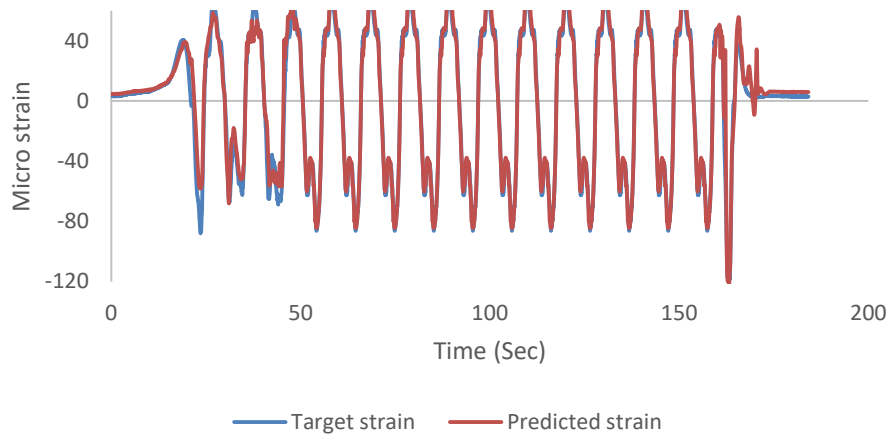

Fig. 7 Target and predicted strain time history at location 5 for Test train at 5 kmph (Stacked LSTM with 80 neurons followed by another 60 neurons, window size=50)

### 5.4 Prediction Case-4

In all the above-mentioned three cases involving test trains, wheel loads of engine and wagons are not much different as may be evident from Fig.3 (g-h). Next, it is tried to utilize strain time history data from bridge members due to passenger train induced vibration, where engine wheel loads are much higher than those of wagons. The pattern of strain time history corresponding to locations 1 and 4[Fig. 3(g-h)] are thus quite different. The strain time history corresponding to location 4 is predicted using the strain time history corresponding to location 1 and very good agreement is observed between predicted strain time history and ground truth as shown in Fig 8.

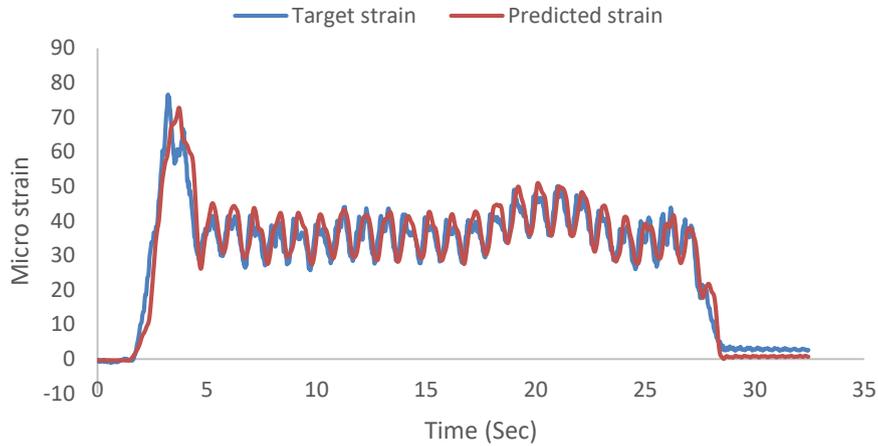

Fig. 8 Target and predicted strain time history at location 4 for passenger train (Stacked LSTM with 80 neurons followed by 60 neurons, window size=60)

Thus, in all the cases, strains in member "1" have only been used for training and strain in many other members are predicted. Table 1 shows the values of RMSE and % Accuracy achieved in prediction using LSTM. It may be noted that all the cases studied here are done using field measured data, which is always mixed with lots of noise and has built in uncertainty. However, the level of accuracy attained using LSTM in prediction based on field measured data is noteworthy. This has huge practical implications. If a network can be trained with already existing data representing strain time histories in important bridge members, the strain time history obtained subsequently from one identified bridge member can be cost-effectively utilized to extract the state of strain in other important bridge members. These strain data can be effectively used to monitor the health of the bridge in future using measured data from only one strain gauge fitted in member "1" for example.

Table 1 Details of accuracy in prediction using LSTM

| Case | RMSE | % Accuracy |
|---|---|---|
| Case 1 (Fig. 4) | 8.929 | 95.19 |
| Case 2 (Fig.5) | 9.361 | 94.66 |
| Case 3 (Fig. 6) | 7.326 | 88.71 |
| Case 3 (Fig. 7) | 7.027 | 84.66 |
| Case 4 (Fig.8) | 4.451 | 86.96 |

## 6      Conclusion

A successful attempt is made for predicting strains in bridge members using LSTM, utilizing actual data from the bridge. The actual data were collected while trains of different axle loads were passing over the bridge at different speeds. While the collection of such data are very important to monitor the health of the bridge, the process is expensive and much more time consuming than training a LSTM model. Thus, introduction of LSTM and successful prediction of strain will certainly open up opportunities for engineers to carry out similar exercise for structural health monitoring in a more efficient and cost-effective manner.

# Acknowledgment

The authors gratefully acknowledges the support of Department of Civil Engineering, IIT Guwahati for providing access to the test data of Bridge no. 40 under Alipurduar Division over river Tista. The bridge data are corresponding to project no. CE/C/AD/225 entitled "Instrumentation of bridges for running of CC+6+2 Tonne loaded BOXN Wagons train on NF Railway" in Civil Engg. Deptt., IIT Guwahati.